\documentclass{article}

\usepackage{microtype}
\usepackage{graphicx}
\usepackage{subcaption}
\usepackage{booktabs} 
\usepackage{svg}
\usepackage{multirow}
 \usepackage{algorithm} 
 \usepackage{algorithmic}
\usepackage{hyperref}
\renewcommand{\algorithmicrequire}{ \textbf{Input:}}
\renewcommand{\algorithmicensure}{ \textbf{Output:}}

\usepackage[preprint]{icml2026}

\usepackage{amsmath}
\usepackage{amssymb}
\usepackage{mathtools}
\usepackage{amsthm}

\usepackage[capitalize,noabbrev]{cleveref}

\theoremstyle{plain}

\theoremstyle{definition}

\theoremstyle{remark}

\usepackage[textsize=tiny]{todonotes}

\icmltitlerunning{One-shot Optimized Steering Vector for Hallucination Mitigation for VLMs}

\begin{document}

\twocolumn[
  \icmltitle{One-shot Optimized Steering Vector for Hallucination Mitigation for VLMs}



  \icmlsetsymbol{equal}{*}

  \begin{icmlauthorlist}
    \icmlauthor{Youxu Shi}{yyy}
    \icmlauthor{Suorong Yang*}{comp}
    \icmlauthor{Dong Liu*}{yyy}
  \end{icmlauthorlist}

  \icmlaffiliation{yyy}{University of Science and Technology of China}
  \icmlaffiliation{comp}{Nanjing University}

  \icmlcorrespondingauthor{Youxu Shi}{syx123@mail.ustc.edu.cn}
  \icmlcorrespondingauthor{Suorong Yang}{sryang@smail.nju.edu.cn}
  \icmlcorrespondingauthor{Dong Liu}{dongeliu@ustc.edu.cn}
  \icmlkeywords{Machine Learning, ICML}

  \vskip 0.3in
]



\printAffiliationsAndNotice{}  

\begin{abstract}
Vision Language Models (VLMs) achieve strong performance on multimodal tasks but still suffer from hallucination and safety-related failures that persist even at scale. Steering offers a lightweight technique to improve model performance. However, steering, whether input-dependent or input-independent, achieves a meaningful trade-off between efficiency and effectiveness. In this work, we observe that steering vectors can generalize across inputs when tasks share aligned semantic intent. Based on this insight, we propose \textbf{OSGA} (\textbf{O}ne-shot \textbf{S}teering with \textbf{G}enerative \textbf{A}nchor), an input-independent framework that improves model performance with a single optimization instance. OSGA first selects an informative sample via a variance-based data selection strategy and learns a single steering vector with a contrastive objective with generative anchor regularization. The resulting vector can be universally applied at a certain layer during inference time without modifying model parameters. Experiments across multiple benchmarks show that a single OSGA-optimized steering vector consistently improves hallucination mitigation and safety enhancement with negligible overhead, highlighting one-shot steering as a practical and scalable solution for reliable VLMs. 
\end{abstract}

\section{Introduction}
\label{introduction}
Recently, Vision Language Models (VLMs)~\citep{vteam2025glm45vglm41vthinkingversatilemultimodal, wang2025internvl35advancingopensourcemultimodal, Qwen3-VL} have demonstrated impressive capabilities across a wide range of complex tasks, including image captioning, visual question answering, and multimodal reasoning.
Despite these advances, hallucinations~\citep{liu2024surveyhallucinationlargevisionlanguage, rani2024visualhallucinationdefinitionquantification, ho2025reviewhallucinationunderstandinglarge} and safety-related issues~\citep{SafetySurvey, jeong2025playingfooljailbreakingllms, liu2024safetyalignmentvisionlanguage, wang2025comprehensivesurveyllmagentstack} remain fundamental obstacles to their reliable deployment, often manifesting as responses that contradict the visual evidence or produce offensive, biased, or misleading content.
Crucially, these issues persist even as models scale~\citep{wang2025alignenoughmultimodaluniversal, jiang2024textttmodscanmeasuringstereotypicalbias}, indicating that simply increasing model capacity or training data is insufficient to address these risks.

To mitigate hallucinations in VLMs, existing methods can be broadly categorized into training-based and training-free approaches.
Training-based methods include supervised fine-tuning (SFT)~\citep{sun2025mitigatinglowlevelvisualhallucinations, li2025analyzingmitigatingobjecthallucination, chen2025perturbollava}, and reinforcement learning from human feedback (RLHF)~\citep{compagnoni2025mitigatinghallucinationsmultimodalllms, yang2025mitigatinghallucinationslargevisionlanguage, yu2024rlaifv}, which can introduce substantial training costs.
Without incurring expensive training costs, another line of work focuses on lightweight and training-free strategies, such as inference-time interventions~\citep{zou2025looktwiceanswermemoryspace, leng2023mitigatingobjecthallucinationslarge, park2024conviscontrastivedecodinghallucination}, post-hoc revision~\citep{Yin_2024, ge2024visualfactcheckerenabling}, and hidden-state editing~\cite{jiang2025interpreting}.
Among these, model steering~\citep{liu2024reducinghallucinationsvisionlanguagemodels, parekh2025learningsteerinputdependentsteering, sivakumar2025steervlmrobustmodelcontrol} emerges as a promising direction due to its simplicity and efficiency. 
By injecting a steering vector into the model's hidden representations~\cite{mikolov2013efficientestimationwordrepresentations}, they directly guide the model's outputs toward a desired semantic direction at inference time without relying on retraining.


While model steering techniques have been widely employed in large language models (LLMs), their applications to VLMs remain largely underexplored, in part due to the complexity of multimodal representations and visual grounding.
For instance, steering a VLM to suppress hallucinated objects may induce overly conservative generative behavior, leading the model to excessively suppress content generation to avoid potential errors.
Moreover, the characteristics of visual tasks can vary across different inputs in terms of complexity and the visual concepts.
Therefore, existing methods~\citep{parekh2025learningsteerinputdependentsteering, shi2025exposinghallucinationssuppressthem, sivakumar2025steervlmrobustmodelcontrol} compute input-specific steering vectors, a form of input-dependent steering, which adjusts the model based on the specific input.
While effective, these approaches introduce non-negligible inference overhead.
In contrast, input-independent approaches~\citep{liu2024reducinghallucinationsvisionlanguagemodels, khayatan2025analyzingfinetuningrepresentationshift} compute a single steering vector applied to all inputs, offering a more efficient and lightweight solution, but with somewhat limited effectiveness.
As a result, a critical question arises: \textbf{How to bridge this gap between the two poses a meaningful trade-off and remains an open research question?}

\begin{table}[t]
\centering
\caption{Comparisons between sample-dependent and sample-independent steering using the CHAIR metric. 
}
\resizebox{\columnwidth}{!}{
\begin{tabular}{l|cccccc}
\toprule
Methods & CHAIR$_S$ $\downarrow$ & CHAIR$_I$ $\downarrow$ & Average $\downarrow$ & Recall $\uparrow$\\
\midrule
Baseline & 53.0 & 14.0 & 33.50 & \textbf{81.00}\\
Sample-dependent & \textbf{47.8} & \textbf{12.7} & \textbf{30.25} & 80.10\\
Sample-independent & 51.4 $\pm$ 3.11 & 13.5 $\pm$ 1.92 & 31.98 $\pm$ 1.63 &  80.11 $\pm$ 1.12\\
\bottomrule
\end{tabular}
}
\label{tab:intro_explore}
\vspace{-3mm}
\end{table}

We observe that steering vectors exhibit cross-input transferability when applied to tasks that share aligned semantic objectives and response patterns. 
Although steering vectors are typically constructed in an input-dependent manner, i.e., generating a steering vector for each sample, their influence on model behavior is not strictly confined to the original input instance.
For instance, a steering vector optimized from a single image-caption pair for hallucination mitigation can remain effective when applied to other image-caption instances.
As shown in Table~\ref{tab:intro_explore}, while sample-specific steering achieves the strongest hallucination suppression, a single steering vector derived from one randomly selected image generalizes well to 500 additional images and consistently mitigates hallucinations.
Notably, computing steering vectors on a per-sample basis incurs substantial inference overhead.
Taken together, these observations suggest that input-dependent steering vectors may implicitly encode task-level or semantic-level control directions that transcend individual samples.
Such behavior points to the existence of a latent input-independent component embedded within ostensibly instance-specific steering signals, motivating a shift from per-sample steering toward more shared and reusable control mechanisms.

In this paper, we propose OSGA (One-shot Steering with Generative Anchor), a lightweight framework for steering VLMs using only a single training instance.
The core idea is to learn a compact steering vector that can generalize across different unseen inputs, enabling efficient and scalable inference-time model steering.
To this end, OSGA first identifies a highly informative anchor example via a variance-based selection strategy, which prioritizes data points that induce diverse and discriminative model responses.
Based on this anchor, we optimize a steering vector that is injected into the decoder layers of the VLM via a contrastive learning objective.
To further enhance optimization stability, we introduce a generative anchor regularization term that constrains the learned steering vector toward a semantically meaningful reference derived from the model's own generative behavior.
Once optimized, the steering vector can be applied universally at inference time without modifying model parameters or requiring additional per-input computation.
This design enables efficient, stable, and effective performance improvements.

Extensive experiments across diverse VLM architectures and benchmarks demonstrate that OSGA consistently improves performance on hallucination mitigation and safety enhancement tasks. Remarkably, a single optimized steering vector generalizes well across unseen inputs and sub-tasks, achieving a favorable trade-off between faithfulness, safety, and task performance. These results highlight the untapped potential of one-shot optimization as a practical and scalable tool for improving the reliability of VLMs.
The contributions are as follows: 
\textbf{(1)} To the best of our knowledge, we are the first to explore one-shot optimized steering in VLMs, demonstrating that hallucinations can be effectively mitigated by introducing only a single token-level steering vector (4096 dimensions), incurring negligible parameter overhead.
\textbf{(2)} We propose OSGA, a one-shot, input-independent steering framework that learns a generalizable steering vector from a single, automatically selected anchor datum, enabling efficient and retraining-free hallucination mitigation for VLMs.
\textbf{(3)} Extensive experiments across various VLM architectures and benchmarks demonstrate the robustness and generality of our method, consistently improving hallucination mitigation and safety performance.

\section{Related work}
\label{related_work}
\subsection{VLMs Hallucination and Safety}
Contemporary VLMs~\citep{bai2025qwen25vltechnicalreport, yang2025qwen3technicalreport, wang2025internvl3_5, zhu2025internvl3} face two primary challenges: hallucination and safety. Whether a VLM can perform well on these two tasks is crucial for determining its reliability and practical deployment. For VLMs, hallucination generally refers to misalignment between the model's response and the visual content, often manifesting as inconsistencies in objects, attributes, and relations~\cite{bai2025hallucinationmultimodallargelanguage}. Safety problems cover a broad range of concerns, referring to the need to mitigate misleading, biased, harmful, or discriminatory outputs, as well as to avoid generating sensitive information that may be inferred from the image. Training-based methods such as Supervised Fine-tuning (SFT)~\cite{chen2025perturbollava}, and reinforcement learning (RL)~\citep{yu2024rlaifv, yu2023rlhf} can effectively address these problems, but they require substantial computational cost and large amounts of training data, making them difficult to adopt and reproduce. More lightweight methods typically include hidden representation editing\cite{jiang2025interpretingeditingvisionlanguagerepresentations, wang2025steeringawayharmadaptive}, alternative decoding strategies\citep{wang-etal-2024-mitigating, chuang2024doladecodingcontrastinglayers}, attention enhancement~\cite{yang2025mitigating} and post-hoc revision~\citep{ge2024visualfactcheckerenabling, sun2023aligninglargemultimodalmodels}. Moreover, the work~\cite{shi2025exposinghallucinationssuppressthem} uses the T2I model to generate the anchor for each input to edit the hidden representations. In contrast, in this work, we propose an optimization method based on a generative anchor that learns an input-independent steering vector using only a single optimization sample. By injecting this vector into a decoder layer of the LLM, we achieve efficient and effective performance improvements.
\subsection{One-shot Optimization/Learning}
Recent efforts in lightweight model adaption increasingly explore \emph{one-shot} optimization for both LLMs and VLMs. Unlike conventional SFT- or RL-based techniques that require extensive datasets and large training costs, one-shot methods aim to derive effective model updates from a single supervision instance. EM~\cite{gao2025oneshotentropyminimization} demonstrates the power of certain unlabeled data points, which are entropy-sensitive. The work~\cite{hossain2025poweronesingleexample} shows the power of a single data point on open-vocabulary segmentation in VLMs with an unsupervised entropy-based layer selection and ensemble strategy. In the safety field, \cite{dunefsky2025oneshotoptimizedsteeringvectors} shows that the steering vectors optimized on a single example are effective in modulating safety-relevant behaviors. In addition, NUGGETS~\cite{li2024oneshotlearninginstructiondata} further demonstrates that the model trained on a small, high-quality subset selected by a single data point can outperform training on the full, large dataset.

\begin{figure*}[htbp]
    \centering
    \includegraphics[width=0.8\linewidth]{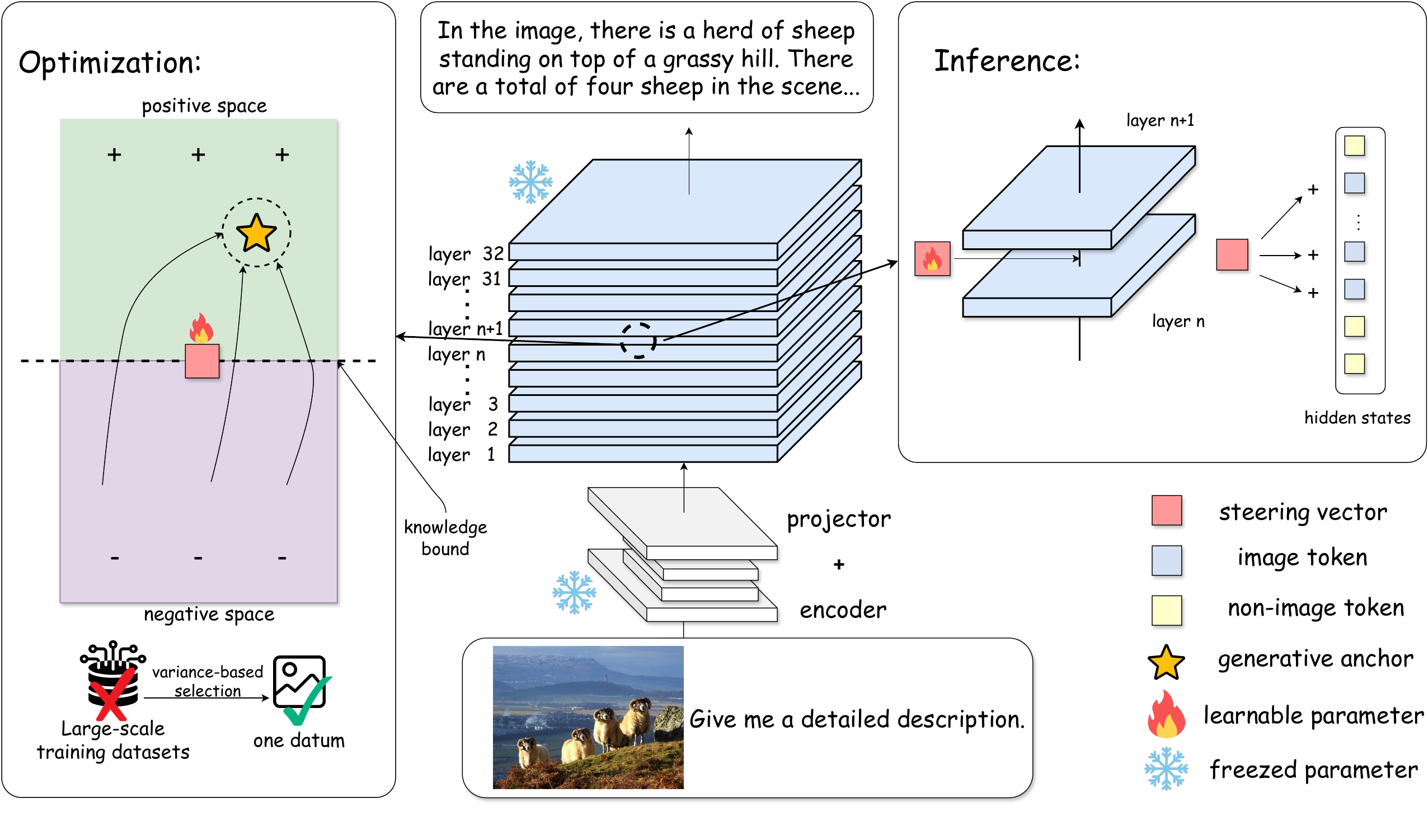}
    \caption{Overview of OSGA: we first select one informative datum from a large-scale training dataset. During the optimization phase (\textbf{left}), we initialize a token randomly as the steering vector, which lies near the knowledge bound. Intuitively, the farther a datum is from the knowledge boundary, the higher the model’s confidence (consistently producing either correct or incorrect answers). In contrast, data points near this boundary represent the most uncertain cases for the model and thus offer substantial potential for optimization. The arrow and the dashed circle surrounding the generative anchor indicate that our optimization objective is to find the optimal point within a soft constraint around the anchor. At the inference stage (\textbf{right}), the learned steering vector is universally added to the image tokens in the hidden states of a selected pair of neighboring layers.}
    \label{fig:overview}
\end{figure*}
\subsection{LLM/VLM Steering}
In general, the steering technique in LLM/VLM focuses on how to find the steering vector. Steering in LLM has largely centered on \emph{contrastive methods}, where the steering vector is derived by contrasting two sets of representations. The steering vectors often represent the difference between such contrastive representations, prompts. In many works, steering vectors are static, which is input-independent. Steering in VLM is still underexplored. VTI~\cite{liu2024reducinghallucinationsvisionlanguagemodels} steers both the image encoder and text decoder with direction between hallucinated and non-hallucinated representations identified by PCA. \cite{li2025internalactivationrevisionsafeguarding} steers both residual streams and selected attention heads, with interventions determined by safety probes. To find an input-independent steering vector, L2S~\cite{parekh2025learningsteerinputdependentsteering} and SteerVLM~\cite{sivakumar2025steervlmrobustmodelcontrol}  train a small auxiliary module to predict the input-specific steering vector. Many studies have shown that steering techniques yield substantial improvements on specific tasks such as safety and open-vocabulary segmentation. However, their applicability to more general tasks, such as hallucination mitigation, remains underexplored and lacks comprehensive evaluation.
\section{Methodology}
\label{method}
\subsection{Preliminaries}
\label{Preliminaries}
\noindent \textbf{Vision Language Model.} Recent VLMs follow a common architecture that combines 
a visual encoder $f_V$, a connector $C$, and an autoregressive language model $f_{\mathrm{LM}}$. 
Following~\cite{parekh2024conceptbasedexplainabilityframeworklarge},
given an input $X=(I,T)$ with image $I$ and text instruction $T$, the model predicts
\begin{equation}
\hat{y} = f(X) = f(I,T).
\end{equation}

The image is encoded into $N_V$ visual tokens 
$h^1,\ldots,h^{N_V} = C\!\circ\! f_V(I)$, 
while the text instruction produces $N_T$ tokens 
$h^{N_V+1},\ldots,h^{N_V+N_T}=\mathrm{Emb}(T)$. 
During generation, previously generated tokens 
$h^p=\mathrm{Emb}(\hat{y}_p)$ for $p>N_V+N_T$ 
are appended autoregressively, and the language model predicts the next token:
\begin{equation}
\hat{y}_{p+1}
= f_{\mathrm{LM}}\big(h^1,\ldots,h^{N_V+N_T},h^{N_V+N_T+1},\ldots,h^p\big).
\end{equation}

Let $h^p_\ell(X)\!\in\!\mathbb{R}^D$ denote the hidden state of the $p$-th token in layer $\ell$. 
A standard transformer stack of $L$ layers updates representations via
\begin{equation}
h^p_{\ell+1}(X)
= h^p_\ell(X)
+ \mathrm{TransformerLayer}_\ell\!\big(h^p_\ell(X)\big),
\end{equation}
where each layer applies self-attention and a feedforward network.

\noindent \textbf{Steering Vector.} Steering vectors provide a lightweight way to modify a VLM's internal representations between decoder layers of LLM without retraining. Given a hidden state $h_\ell \in \mathbb{R}^d$ at layer $\ell$, a steering vector \textbf{$v$} is an offset computed from contrastive signals, and is injected into the forward pass as
\begin{equation}
\tilde{h}_{\ell} = h_{\ell} + \alpha v,
\end{equation}
where $\alpha$ controls the strength of the intervention. During decoding, this shift representation $\tilde{h}_{\ell}$ guides the model toward desired behaviors, such as reducing hallucinations or enforcing safety, while preserving most of the original semantics.

\noindent \textbf{Generative Anchor.} 
 As a contrastive signal, the generative anchor~\cite{shi2025exposinghallucinationssuppressthem} is computed by feeding the model's initial hallucination-prone output to the T2I (Text-to-Image) model to reconstruct the image, and then extracting its visual embedding as an "anchor". The core editing is as follows:
\begin{equation}
\begin{aligned}
\tilde{h}_{k, \ell}
&= h_{k, \ell} + \alpha ~C\!\circ\! f_V(I)-\beta ~C\!\circ\! f_V(I^\prime),\\
&\qquad k\in \mathcal{H}_{img},\; \ell \in [1, L]
\end{aligned}
\end{equation}
where $h_{k,\ell}$ stands for the embedding of the $k$-th token at the $\ell$-th decoder layer, $\mathcal{H}_{img}=\{i_1, i_1+1, \dots i_n\}$ represents the set of positions corresponding to the image-related tokens, $\alpha$ and $\beta$ are scalar coefficients controlling the contributions from the original image $I$ and the reconstructed image $I^\prime$, respectively. 
\subsection{Overview}
As illustrated in Figure \ref{fig:overview}, One-shot Steering with Generative Anchor (OSGA) begins by identifying an informative one-shot datapoint using a variance-based data selection strategy. This datapoint is used to optimize a lightweight steering vector interleaved in the decoder layer of VLM. Specifically, we choose a decoder layer $\ell$ in the VLM and introduce a trainable vector, which can be viewed as a learnable token injected into that layer. We then optimize this vector using the proposed training objective, producing a task-specific steering vector that captures the semantics around the model's knowledge boundary. The optimized vector is subsequently added to the hidden representations of all image tokens at layer $\ell$.
Empirically, as shown in Table \ref{tab:intro_explore}, this learned ‘token-level’ steering mechanism may generalize across different inputs well: a single optimized vector yields consistent improvements across unseen inputs, demonstrating both robustness and strong cross-instance transferability. 

\subsection{One-shot Steering with Generative Anchor}
\label{OSGA}

\noindent \textbf{Optimization Objective.}
Inspired by \cite{dunefsky2025oneshotoptimizedsteeringvectors}, we adopt a contrastive steering (promotion and suppression) method, which is a mixed steering. Considering that a single training datapoint may not be sufficient to reach an optimal solution within a limited number of optimization steps, we introduce a generative anchor. Extensive experiments show that steering with this anchor yields substantially better performance of VLMs. Therefore, we incorporate the generative anchor into the loss function as a regularization term, treating it as a soft constraint that guides the optimization trajectory toward a more promising region. We compute the generative anchor as follows:
\begin{equation}
E_{GA}=\alpha ~C\!\circ\! f_V(I)-\beta ~C\!\circ\! f_V(I^\prime)
\end{equation}
where the $I^\prime$ is the reconstructed image using captions from VLM itself.

We use $y = \{y_1, \ldots, y_i\}$ to denote the sequence of tokens generated by the model after the \(i\)-th autoregressive step, \(x\) represents the input image and prompt, and \(v\) denotes the steering vector. Furthermore,
$P_{\text{model}}(y \mid x ~; v)$
denotes the probability assigned by the model to the sequence \(y\) given the prompt \(x\) when the steering vector \(v\).

For promotion steering, where the $x$ stands for positive data:
\begin{equation}
\mathcal{L}_{+}(x, y; v)
= \begin{aligned}[t]
&- \sum_{k=0}^{m-1} \log P_{\text{model}}\!\left(
y_{k+1} \mid y_k, \ldots, y_1, x; v
\right) \\
&\quad + \gamma\, \mathrm{Dis}(v, E_{GA}) .
\end{aligned}
\end{equation}
For suppression steering, $x$ stands for negative data:
\begin{equation}
\mathcal{L}_{-}(x, y; v)
= \begin{aligned}[t]
&- \sum_{k=0}^{m-1} \log (1- P_{\text{model}}\!\left(
y_{k+1} \mid y_k, \ldots, y_1, x; v
\right)) \\
&\quad + \gamma\, \mathrm{Dis}(v, E_{GA}) .
\end{aligned}
\end{equation}

Here, $\gamma$ serves as a relaxation coefficient for the soft constraint in the optimization objective. It is introduced to balance the scale between the cross-entropy loss and the constraint penalty term, preventing the latter from dominating the overall optimization due to magnitude mismatch. In practice, $\gamma$ can be adaptively set as:
\begin{equation}
\gamma = \frac{\mathcal{L}_{\text{cross-entropy}}}{\mathrm{Dis}(v,E_{GA}) + \epsilon},
\end{equation}
where $\epsilon$ denotes a small constant for numerical stability.

\begin{table*}[t]
\centering
\caption{Performance comparisons on POPE across different settings and datasets on LLaVA-v1.5.}
\label{tab:pope}
\resizebox{.95\textwidth}{!}{
\begin{tabular}{@{}clcccccccccc@{}}
\toprule
\multirow{2}{*}{Dataset} & \multirow{2}{*}{Methods} & \multicolumn{2}{c}{Random} & \multicolumn{2}{c}{Popular} & \multicolumn{2}{c}{Adversarial} & \multicolumn{2}{c}{Average} \\
\cmidrule(lr){3-4} \cmidrule(lr){5-6} \cmidrule(lr){7-8} \cmidrule(lr){9-10}
& & Accuracy $\uparrow$ & F1-score $\uparrow$ & Accuracy $\uparrow$ & F1-score $\uparrow$ & Accuracy $\uparrow$ & F1-score $\uparrow$ & Accuracy $\uparrow$ & F1-score $\uparrow$ \\
\midrule
\multirow{5}{*}{MSCOCO}
& Baseline  & 83.49 & 82.28 & 79.98 & 79.34 & 76.03 & 76.26 & 79.83 & 79.29 \\
& ICD   & 84.87 & 83.27 & 82.93 & 81.45 & 81.07 & 79.96 & 82.96 & 81.56 \\
& VCD   & 86.84 & 86.83 & 82.65 & 83.37 & 77.31 & 79.28 & 82.27 & 83.16 \\
& OPERA   & 87.53  & 86.45 & 84.21  & 83.50 & 80.88 & 80.69 & 84.21 & 83.55 \\
& Ours & \textbf{87.87} & \textbf{86.43} & \textbf{86.37} & \textbf{85.00} & \textbf{84.50} & \textbf{83.29} & \textbf{86.29} & \textbf{84.91} \\
\cmidrule{1-10}
\multirow{5}{*}{A-OKVQA}
& Baseline  & 83.45 & 82.56 & 79.90 & 79.59 & 74.04 & 75.15 & 79.13 & 79.10 \\
& ICD   & 85.57 & 85.06 & 81.93 & 81.95 & 77.43 & 78.99 & 81.64 & 82.00 \\
& VCD   & 86.15 & 86.34 & 81.85 & 82.82 & 74.97 & 77.73 & 80.99 & 82.30 \\
& OPERA   & 88.27 & 87.54 & 85.17 & 84.74 & 79.37 & 79.97 & 84.27 & 84.08 \\
& Ours & \textbf{90.13} &  \textbf{89.75}  &  \textbf{85.27}  &  \textbf{85.43}  &   \textbf{79.90}  & \textbf{78.90}   &  \textbf{85.10}  &  \textbf{84.69}   \\
\cmidrule{1-10}
\multirow{5}{*}{GQA}
& Baseline  & 83.73 & 82.95 & 78.17 & 78.37 &75.08  & 76.06 & 78.99 & 79.13 \\
& ICD   & 84.90 & 84.22 & 78.37 & 78.81 & 75.97 & 76.93 & 79.75 & 79.99 \\
& VCD   & 86.65 & 86.99 & 80.73 & 82.24 & 76.09 & 78.78 & 81.16 & 82.67 \\
& OPERA   & 83.73 & 82.95 & 78.17 & 78.37 & 75.08 & 76.06 & 78.99 & 79.13 \\
& Ours &  \textbf{89.03}  & \textbf{88.52}   &  \textbf{80.80}  & \textbf{81.60}   &  \textbf{76.87}  &  \textbf{78.53}  &  \textbf{82.23}  &  \textbf{82.88}  \\
\bottomrule
\end{tabular}}
\end{table*}
All optimization data used in this process is obtained directly from the model itself. Therefore, the entire optimization can be regarded as a self-supervised procedure, without relying on any manually curated or specially designed data.
Algorithm~\ref{alg:one_shot_steering} presents the detailed optimization process.
\begin{figure}[]
    \centering
    \includegraphics[width=.95\linewidth]{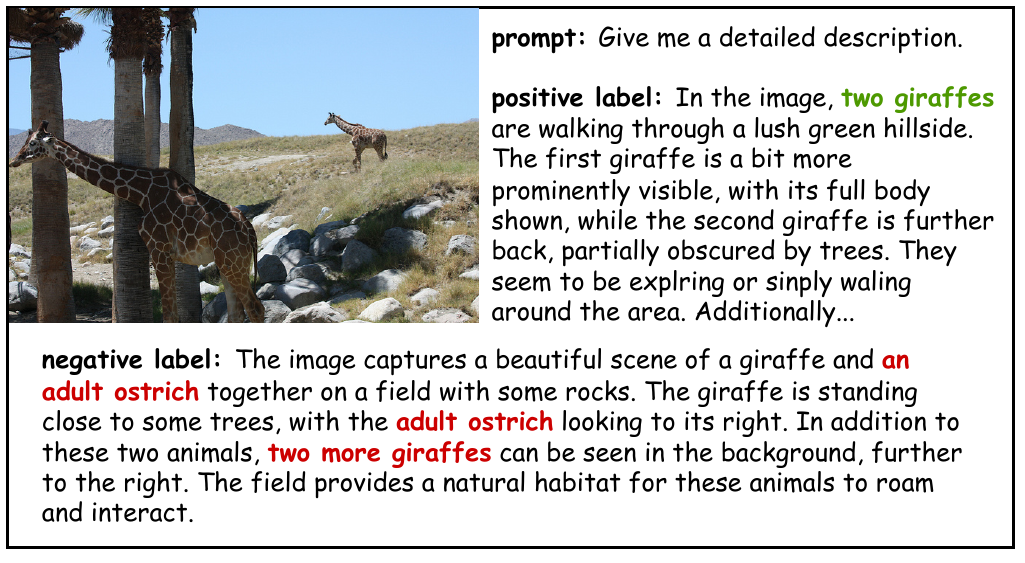}
    \caption{Illustration of a data point selected by the data selection strategy. Both positive and negative labels are generated by the model itself, making the selection process self-supervised.}
    \label{fig:oneshot_example}
\end{figure}
\subsection{Data Selection}
\label{Data_selection}
The motivation of OSGA is to investigate the effectiveness of one-shot data for optimizing the steering vector in VLMs.
Prior work \citep{razin2025makesrewardmodelgood, wang2025reinforcementlearningreasoninglarge,yang2024clip} shows that models tend to give highly deterministic answers to certain prompts, i.e., consistently correct or consistently wrong.
Instead of randomly selecting a data point, we identify a data point that lies near the model's knowledge boundary, enabling more effective optimization.

\begin{algorithm}[t]
\caption{The Algorithm of OSGA.}
\label{alg:one_shot_steering}
\begin{algorithmic}[1]
\REQUIRE data point $x^\ast = \{(I^1, q, a^1), (I^2, q, a^2)\}$, VLM $\mathcal{M}$, target decoder layer $\ell$, optimization steps $T$, learning rate $\eta$, generative Anchor $E_{GA}$, steering factor $\alpha$. 
\ENSURE Optimized steering vector $v$, and steered hidden representation $\tilde{h}_\ell$
\STATE \textbf{Initialization:} $v \in \mathbb{R}^{1 \times d_{\text{model}}}$, freeze $\mathcal{M}$ \textbackslash $v$.
\FOR{$t = 1$ to $T$}
    \STATE $\mathcal{L}(x; v) \leftarrow \mathcal{L}_{+}(x, y; v) + \mathcal{L}_{-}(x, y; v)$
    \STATE $v \leftarrow v - \eta \nabla \mathcal{L}(x; v)$
\ENDFOR
\STATE $h_\ell \leftarrow \mathcal{M}_\ell(x^\ast)$ \{extract hidden states from layer $\ell$\}
\FOR{$k \in \mathcal{H}_{img}$}
    \STATE $\tilde{h}_{k,\ell} \leftarrow h_{k,\ell} + \alpha v$
\ENDFOR
\end{algorithmic}
\end{algorithm}
Following the approach in \cite{gao2025oneshotentropyminimization}, we adopt a variance-based data selection strategy to select such a data point.
We begin by randomly selecting a subset $\mathcal{D}$ of Visual Question Answering (VQA) data. For each VQA pair, we roll out $k$ independent samples from the model:
\begin{equation}
\mathcal{Y}(x) = \left\{ y^{(1)}, y^{(2)}, \ldots, y^{(k)} \right\}, 
\qquad y^{(i)} \sim p_\theta(\cdot \mid x).
\end{equation}
Then we compute the output score $\mathcal{R}(x)$ as:
\begin{equation}
\mathcal{R}(x) 
= \frac{1}{k} \sum_{i=1}^{k} \mathcal{S}(y^{(i)}),
\end{equation}
where $\mathcal{S}(\cdot)$ denotes the score function for each sample, and $\mathcal{R}(\cdot)$ represents the average score over the $k$ samples.\
Next, we can further compute the variance of the scores corresponding to these samples:
\begin{equation}
Var(x)=\frac{1}{k} \sum_{i=1}^{k}(\mathcal{S}(y^{(i)})-R(x))^2
\end{equation}
Such variance serves as a measure of the model's confidence for this visual question: A lower variance indicates higher confidence, whether consistently correct or wrong (meaning inside or outside the model’s knowledge), and suggests that the question sample is far from the knowledge boundary, which contains less information gain for optimization. So the objective of data selection, which aims to identify the questions that maximize the variance, is defined as follows:
\begin{equation}
x^\star=\arg \max_{x \in \mathcal{D}} \operatorname{Var}(x),
\end{equation}
where $\mathcal{D}$ denotes the data subset pool.\
For different tasks, the score function should vary due to the gap in data characteristics. For the hallucination task, we define $\mathcal{S}(x) = 1-h(x)$, where $h$ means the average CHAIR score. For safety tasks, we use prompts with binary success outcomes, in which case $\mathcal{S}(x)$ becomes the indicator function $\mathbb{I}(x)$. Specifically:
\begin{equation}
\begin{aligned}
\mathcal{S}(x) &=
\begin{cases}
1 - h(x), & \text{hallucination task}, \\[6pt]
\mathbb{I}(x), & \text{safety task},
\end{cases} \\[10pt]
h(x) &= \mathrm{CHAIR}.
\end{aligned}
\end{equation}
Under this data selection strategy, we obtain an input condition within a limited subset that best characterizes those lying near the model’s knowledge boundary, which are the cases where the model sometimes answers correctly and sometimes fails. This naturally forms a pair of contrastive data pairs, providing the richest possible supervisory signal for optimizing the steering vector. In this work, we select two one-shot data points for optimizing the steering vector, corresponding respectively to the hallucination task and safety-relevant tasks. 
The hallucination task\footnote{For ethical reasons and to avoid offensive content, safety-related images are omitted deliberately.} is shown in Figure \ref{fig:oneshot_example}.

\section{Experiments}
\label{exp}
\subsection{Experiment Setup}
\label{setup}
\noindent \textbf{Models and Benchmarks.} We compare OSGA with two different model architectures, including  \textbf{LLaVA-v1.5}~\cite{liu2024improvedbaselinesvisualinstruction} that employs linear projection for visual-text alignment, and \textbf{Qwen2-VL}~\cite{Qwen2VL} that has a more complex alignment mechanism.
 In this work, we first evaluate the performance of OSGA on hallucination tasks, covering objective, attributive, and relational hallucinations.
 Thus, the benchmarks include CHAIR~\cite{rohrbach2019objecthallucinationimagecaptioning}, POPE~\cite{li2023evaluatingobjecthallucinationlarge}, MME~\cite{fu2025mmecomprehensiveevaluationbenchmark}, FaithScore~\cite{jing2024faithscorefinegrainedevaluationshallucinations}, HalFscore~\cite{chen2025perturbollava}, and GAVIE~\cite{liu2024mitigatinghallucinationlargemultimodal}.
For safety-related tasks, we also assess the effectiveness of OSGA on GOAT-Bench~\cite{lin2025goatbenchsafetyinsightslarge}. 

\subsection{Results on Object Hallucination}
\label{hallu}
\noindent \textbf{POPE Benchmark.} 
We evaluate OSGA on the POPE benchmark across MSCOCO~\cite{lin2015microsoftcococommonobjects}, A-OKVQA~\cite{schwenk2022aokvqabenchmarkvisualquestion}, and GQA~\cite{hudson2019gqanewdatasetrealworld}, under random, popular, and adversarial query settings.
POPE provides a comprehensive testbed for assessing object hallucination by varying both dataset characteristics and prompt distributions.
As shown in Table~\ref{tab:pope}, OSGA consistently outperforms all competing methods across all datasets, metrics, and evaluation settings, demonstrating strong robustness and generality.
In particular, on MSCOCO, OSGA improves the average Accuracy and F1-score by 6.46\% and 5.62\%, respectively, compared to the baseline, and consistently surpasses prior SOTA methods such as OPERA.
Overall, these results demonstrate the effectiveness of OSGA for mitigating object hallucination in diverse multimodal scenarios, highlighting its robustness across datasets and query settings.

\begin{table}[]
\centering
\caption{Performance of hallucination mitigation on CHAIR using LLaVA-v1.5. GA-w/o stands for the original GA method (using GA without optimization) in~\cite{shi2025exposinghallucinationssuppressthem}.}
\resizebox{\columnwidth}{!}{
\begin{tabular}{l|cccccc}
\toprule
Methods & CHAIR$_S$ $\downarrow$ & CHAIR$_I$ $\downarrow$ & Average $\downarrow$ & Recall $\uparrow$ & F1-score $\uparrow$\\
\midrule
Baseline & 53.0 & 14.0 & 33.50 & \textbf{81.00} & 73.04\\
OPERA & 47.8 & 14.6 & 31.20 & 76.80 & 72.58 \\
ICD  & 56.2 & 16.3 & 36.25 & 16.31 & 25.96\\
VCD  & 48.7 & 14.9 & 31.80 & 77.32 & 72.47\\
MemVR  & 46.6 & 13.0 & 29.80 & 80.80 & 75.13\\
GA-w/o  & 47.8 & 12.7 & 30.25 & 80.10 & 74.60\\
\midrule
RLAIF-V & 18.1 & 4.7 & 11.40 & 59.20 & 71.00\\
\midrule
Ours & \textbf{28.8} & \textbf{7.7} & \textbf{18.25} & 77.30 & \textbf{79.46}\\
\bottomrule
\end{tabular}
}
\label{tab:chair}
\end{table}
\noindent \textbf{CHAIR Benchmark.} 
CHAIR provides fine-grained metrics that jointly measure hallucination severity and recall. 
As shown in Table \ref{tab:chair}, we compare OSGA with various baselines, including training-free methods and fine-tuning-based methods, RLAIF-V.
Among all training-free methods, OSGA achieves substantially lower CHAIR$_S$ and CHAIR$_I$ scores while preserving high recall, indicating a strong ability to suppress hallucinated objects without overly sacrificing correct object mentions.
In contrast to RLAIF-V, while OSGA does not eliminate hallucinations as aggressively as training-based approaches, it maintains a significantly higher recall. This shows a more favorable balance between hallucination suppression and content coverage.
This trade-off is reflected in the F1 score, where OSGA attains the best perfomrance, effectively mitigating hallucinations without sacrificing content information. 



\subsection{Results on Attribute Hallucination}
\noindent \textbf{MME Benchmark.} 
We further evaluate OSGA on the MME benchmark using LLaVA-v1.5.
As shown in Table~\ref{tab:mmehall}, OSGA achieves the best overall performance on MME-Hall, improving the total score from 643.3 to 699.5, corresponding to an 8.74\% gain over the baseline.
Notably, these improvements are consistent across both object-level and attribute-level evaluations. 
At the object level, OSGA outperforms all competing methods on both existence and count, indicating improved recognition of object presence and quantity.
At the attribute level, OSGA yields particularly strong gains on position and color, which are known to be challenging due to their fine-grained nature.
Compared with GA-w/o, which directly applies a generative anchor without optimization, OSGA demonstrates substantially improved performance on attribute-centric tasks such as count and color.
This observation aligns with prior findings that text-to-image models struggle to faithfully reconstruct fine-grained attributes.
In contrast, OSGA's optimized steering procedure enables the learned vector to better capture attribute-level control signals, resulting in more accurate and stable mitigation of attribute hallucination.


\begin{table}[t]
\centering
\caption{Performance comparisons on MME-Hall across different subcategories on LLaVA-v1.5.}
\label{tab:mmehall}
\resizebox{\columnwidth}{!}{
\begin{tabular}{l c | cc | cc}
\toprule
\multirow{2}{*}{Methods} & MME-Hall
& \multicolumn{2}{c|}{Object-Level} 
& \multicolumn{2}{c}{Attribute-Level} \\
\cmidrule(lr){3-4} \cmidrule(lr){5-6}
&Total$\uparrow$ & Existence$\uparrow$ & Count$\uparrow$ 
  & Position$\uparrow$ & Color$\uparrow$ \\
\midrule
Baseline & 643.3 & 190.0 & 155.0 & 128.3 & 170.0 \\
ICD      & 583.3 & 185.0 & 130.0 & 121.7 & 146.7 \\
VCD      & 648.3 & 190.0 & 155.0 & 133.3 & 170.0 \\
OPERA    & 610.0 & 195.0 & 128.3 & 121.7 & 165.0 \\
GA-w/o    & 667.3 & 193.0 & 156.7 & 150.0 & 167.7 \\
Ours        & \textbf{699.5} & \textbf{195.0} & \textbf{165.0} & \textbf{160.0} & \textbf{179.5} \\
\bottomrule
\end{tabular}
}
\end{table}

\begin{figure}[]
    \centering
    \includegraphics[width=1.0\columnwidth]{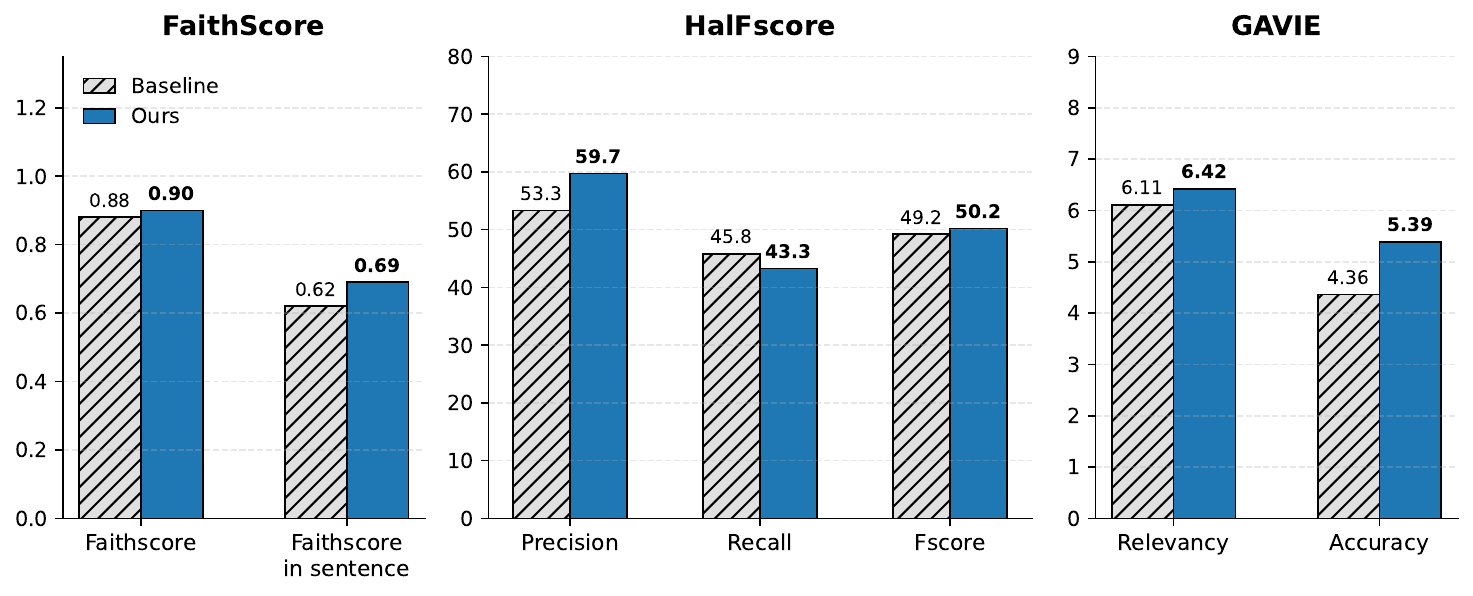}
    \caption{Quantitative comparisons on FaithScore, HaIFscore, and GAVIE using LLaVA-v1.5. The results demonstrate that our method consistently outperforms the baseline across the majority of metrics. }
    \label{fig:diverse}
\end{figure}
\subsection{Results on Relation and General Hallucination}
\noindent \textbf{FaithScore.} FaithScore is a reference-free metric designed to evaluate the factual faithfulness of free-form responses generated by VLMs. 
As shown in Figure~\ref{fig:diverse}, OSGA achieves a notable gain at the subsentence level, indicating a clear reduction in fine-grained, local hallucination. 
This suggests that OSGA improves visual grounding at the sentence level, correcting specific factual errors rather than trivially boosting the score by shortening or simplifying responses.
This indicates that the improvements are not driven by response compression, but by enhanced alignment between visual evidence and textual claims.

\begin{table*}[]
\centering
\caption{Performance comparisons on the GOAT benchmark across different subcategories on LLaVA-v1.5.}
\label{tab:goat}
\resizebox{0.75\textwidth}{!}{
\begin{tabular}{l cc cc cc cc | cc}
\toprule
\multirow{2}{*}{Methods} & \multicolumn{2}{c}{Misogyny} & \multicolumn{2}{c}{Offensiveness} & \multicolumn{2}{c}{Sarcasm} & \multicolumn{2}{c}{Harmfulness} & \multicolumn{2}{c}{Average}\\
\cmidrule(lr){2-3} \cmidrule(lr){4-5} \cmidrule(lr){6-7} \cmidrule(lr){8-9} \cmidrule(lr){10-11}
& Acc. & F1 & Acc. & F1 & Acc. & F1 & Acc. & F1 & Acc. & F1\\
\midrule
Baseline & 49.40 & 32.76 & 41.59 & 30.48 & 50.11 & 33.58 & 44.00 & 34.52 & 46.28 & 32.84\\
Ours   & \textbf{56.90} & \textbf{36.71} & \textbf{47.51} & \textbf{57.61} & \textbf{50.38} & \textbf{38.30} & \textbf{68.02} & \textbf{47.12} & \textbf{55.70} & \textbf{44.94}\\
\bottomrule
\end{tabular}
}
\end{table*}
\begin{figure*}[]
    \centering
    \begin{minipage}{0.49\linewidth}
        \centering
        \includegraphics[width=\linewidth]{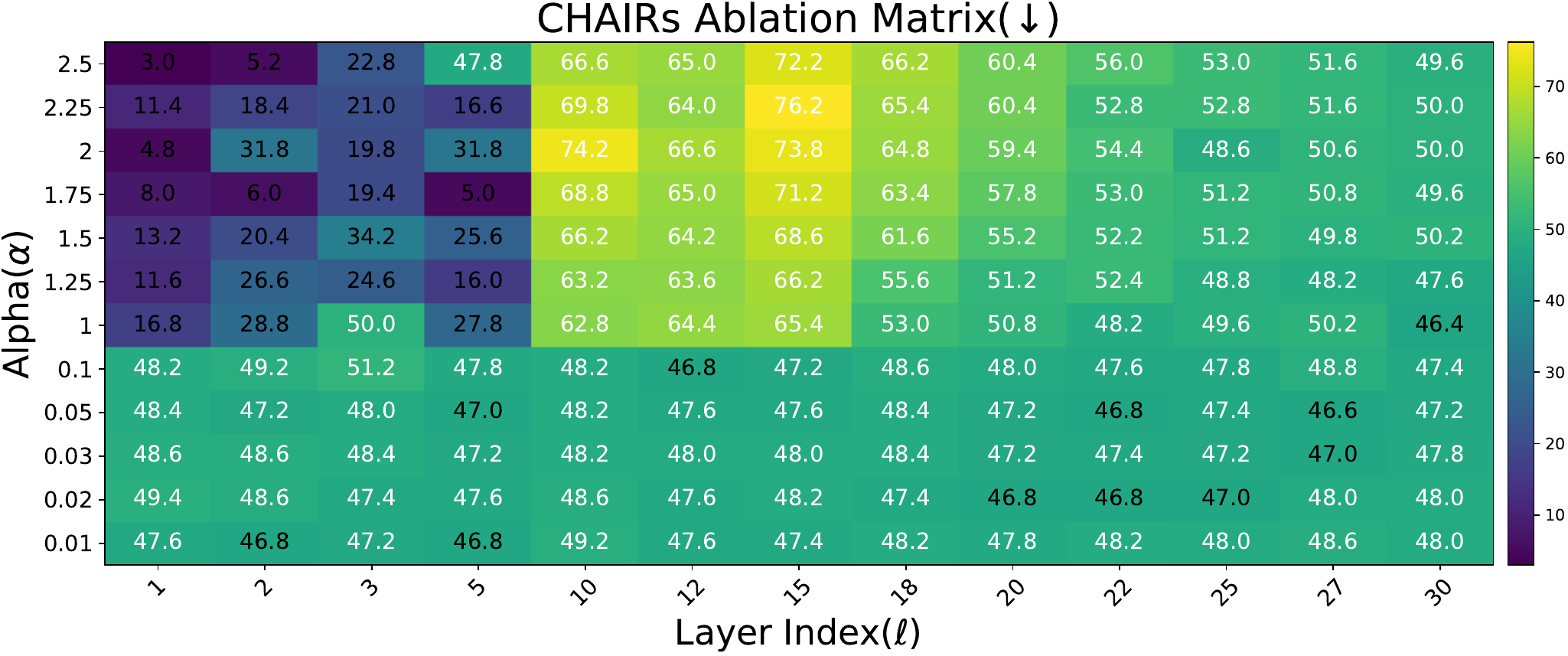}
    \end{minipage}
    \hfill
    \begin{minipage}{0.49\linewidth}
        \centering
        \includegraphics[width=\linewidth]{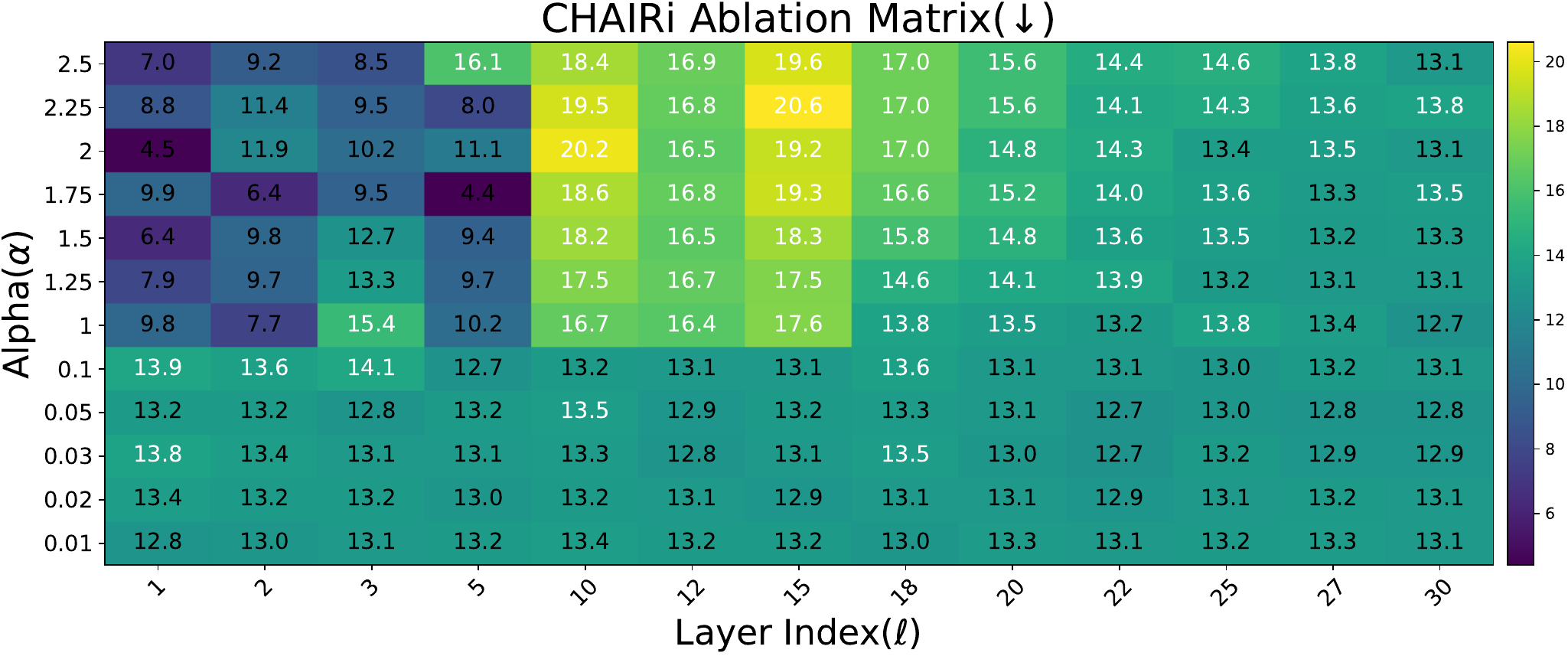}
    \end{minipage}
    \caption{Ablation matrices for OSGA  steering strength ($\alpha$) and Layer $\ell$ on LLaVA-v1.5. A darker color signifies better performance.}
    \label{fig:ablation}
\end{figure*}
\noindent \textbf{HalFscore.} HalFscore evaluates hallucination and omission in dense image captioning. 
As shown in Figure~\ref{fig:diverse} (middle), OSGA achieves a substantial improvement in precision with only a marginal decrease in recall, leading to a clear gain in the overall F1 score. 
This trend indicates that OSGA effectively suppresses hallucinated content while largely preserving correct visual descriptions. 
Importantly, the improvement is not achieved by aggressively filtering or shortening responses; instead, it reflects a more reliable selection of visually grounded information.

\noindent \textbf{GAVIE Benchmark.} GAVIE evaluates hallucinations from two complementary perspectives: \textit{Relevancy}, which evaluates the instruction-following behavior, and \textit{Accuracy}, which assesses the faithfulness of visual content in the generated responses.
As shown in Figure~\ref{fig:diverse} (right), OSGA improves both relevancy and accuracy, indicating that the optimized steering vector enhances not only the model's instruction-following capability but also its ability to faithfully capture visual information.
\subsection{Results on Safety Problem}
\label{safety}
GOAT-Bench evaluates a model’s ability to detect subtle and diverse forms of social abuse in multimodal settings.
As shown in Table~\ref{tab:goat}, OSGA consistently outperforms the baseline across all four categories, yielding substantial improvements in both accuracy and F1 score.
Notably, the gains are particularly pronounced in challenging categories such as harmfulness and offensiveness, where OSGA improves the F1 score by a large margin.
These improvements indicate that the optimized steering vector enhances the model’s sensitivity to safety-critical cues without introducing excessive false positives.



\subsection{Ablation Study}
\label{ablation}

\noindent \textbf{Effect of Data Selection.}
In OSGA, we propose an unsupervised data selection strategy to identify datapoint that can optimize a steering vector with strong generalization ability. A natural question is how our optimization behaves when using randomly selected data. To examine this, we randomly sample 10 datapoints from the original subset and independently perform steering optimization on each.
As shown in Table~\ref{tab:strategy_ablation}, we report the mean and variance results. 
It shows that using the randomly selected data points for optimization not only yields inferior average performance compared with our proposed selection strategy but also exhibits considerably higher instability.
Thus, we validate the effectiveness of our proposed data selection strategy.

\noindent \textbf{Effect of Layer $\ell$.}
Figure~\ref{fig:ablation} presents the performance of our method across different layers on CHAIR$_S$, CHAIR$_I$, and Recall. The performance of steering varies substantially across different layers under different steering strengths. When the steering coefficient $\alpha < 1$, our method consistently and stably mitigates hallucinations across layers. In contrast, when $\alpha > 1$, the hallucination rate generally exhibits a non-monotonic trend, first decreasing and then increasing. Specifically, steering applied to shallow layers (layers 1-5) leads to a pronounced reduction in hallucination, whereas modifying intermediate layers (layers 10-22) results in a significant increase in hallucination. When steering is applied to deeper layers (after layer 25), the hallucination rate is only slightly reduced.

\begin{table}[t]
\centering
\caption{Comparison between our data selection strategy and random selection. The results show that optimization data selected by our strategy leads to more stable performance and achieves a higher F1 score.}
\resizebox{\columnwidth}{!}{
\begin{tabular}{l|cccccc}
\toprule
Methods & CHAIR$_S$ $\downarrow$ & CHAIR$_I$ $\downarrow$ & Average $\downarrow$ & Recall $\uparrow$ & F1-score $\uparrow$\\
\midrule
Baseline & 53.0 & 14.0 & 33.50 & 81.00 & 73.04\\
random & 24.6 $\pm$ 2.04 & 9.4 $\pm$ 1.18 & 16.97 $\pm$ 1.036 & 67.4 $\pm$ 2.729 & 74.40 $\pm$ 1.97 \\
Ours & \textbf{25.8 $\pm$ 0.55} &  \textbf{7.4 $\pm$ 0.56} & \textbf{17.6 $\pm$ 0.56} & \textbf{77.6 $\pm$ 1.01}  & \textbf{79.93 $\pm$ 1.15}\\
\bottomrule
\end{tabular}
}
\label{tab:strategy_ablation}
\end{table}

\noindent \textbf{Effect of Weight Factor $\alpha$.}
As shown in Figure~\ref{fig:ablation}, we vary steering strength $\alpha$ from small to large values and observe distinct behaviors across different layer depths. Overall, under relatively small steering strengths ($\alpha$ in 0.01-0.1), our method consistently and stably reduces hallucinations. However, when the steering strength becomes large ($\alpha>1$), the effects become layer-dependent: at shallow layers (layer 1-5), stronger steering significantly mitigates hallucinations; at intermediate layers (layers 10-22), it instead exacerbates hallucination; and at deeper layers (after layer 25), the hallucination reduction effect becomes relatively weaker.

\noindent \textbf{Hyperparameter Selection Advice.}
By referring to the recall ablation matrix results in the Appendix~\ref{appendix}, we observe a clear trade-off between hallucination reduction and recall. When the steering is small ($\alpha<1$), applying steering to any layer can reduce hallucinations, demonstrating the stability and robustness of our method. In contrast, under larger steering strengths, steering at shallow layers leads to a more pronounced reduction in hallucination. demonstrating the effectiveness of our method again.

\section{Conclusion}
This work investigates the generalization of one-shot steering vectors in VLMs and introduces \textbf{OSGA}, a lightweight steering framework for mitigating hallucination risks.
Unlike prior methods that rely on large training datasets, OSGA optimizes a steering direction from a single informative instance. 
Through the inference-time steering, OSGA effectively reduces hallucinations with high efficiency.
Experiment results demonstrate that the proposed method achieves significant improvements in hallucination mitigation, safety-related tasks, and general multimodal understanding. 
We hope our work inspires further research on hallucination mitigation from a steering perspective and believe OSGA will serve as a valuable tool for the community, enabling 
safer, more reliable, and more scalable deployment of vision-language models in real-world applications.



\section*{Impact Statement}

This work aims to improve the safety and reliability of Vision–Language Models (VLMs), which are increasingly deployed in real-world applications. By focusing on hallucination mitigation and safety-related failures, our method helps reduce responses that contradict visual evidence or generate offensive, biased, or misleading content. In particular, we demonstrate that the proposed approach is effective not only for hallucination-related tasks, but also for safety-critical scenarios such as analyzing meme images that may contain offensiveness, discrimination, or other harmful attributes.

By providing a lightweight and scalable steering mechanism, this work lowers the barrier to improving the trustworthiness of existing VLMs without requiring costly retraining or access to large datasets. We believe this contributes positively to the development of more reliable multimodal AI systems and supports safer deployment in user-facing settings. At the same time, our method is designed as a controllable inference-time intervention, which can be combined with existing safety practices rather than replacing them. Overall, this work highlights the potential of efficient steering techniques as a practical tool for enhancing the alignment, safety, and trustworthiness of multimodal AI systems.


\bibliography{example_paper}
\bibliographystyle{icml2026}

\newpage
\appendix
\onecolumn
\section{Discussion and Future Work}

We discuss some limitations for OSGA. In this work, OSGA shows a significant improvement when it is applied to tasks such as image captioning and open-ended questions. For tasks that require specific answer formats, such as Yes/No questions, the gains of this method are relatively limited, and overfitting can easily occur during optimization, leading the model to consistently respond with “yes” or “no.”

\section{Extensive experimental details}
\label{appendix}
\subsection{CHAIR}
We uniformly adopt the 500 images from an MS-COCO subset sampled in the work \cite{zou2025looktwiceanswermemoryspace} to ensure the results are directly comparable. CHAIR is proposed to evaluate object hallucination in image captioning tasks. It uses two metrics as the evaluation score: per-instance (CHAIR$_I$) and per-sentence (CHAIR$_S$). These are defined as:
\begin{equation}
\begin{aligned}
        \text{CHAIR}_I = \frac{|\{\text{hallucinated objects}\}|}{|\{\text{all objects mentioned}\}|}, \;
     \text{CHAIR}_S = \frac{|\{\text{sentences with hallucinated objects}\}|}{|\{\text{all sentences}\}|}
\end{aligned}.
\end{equation}
As mentioned in Table \ref{tab:chair}, we use the $F_1$ score to evaluate the trade-off performance as follows:
\[
F_{1} = 2 \cdot \frac{\text{Precision} \cdot \text{Recall}}{ \text{Precision} + \text{Recall}}
\]

\[
\text{Precision} = 1 - \frac{CHAIR_S + CHAIR_I}{2}
\]
\subsection{POPE}
We adopt the official benchmark proposed by ~\cite{li2023evaluatingobjecthallucinationlarge}, which consists of three evaluation settings: random, popular, and adversarial, each containing 3000 question-answer pairs. In the random setting, the queried objects are randomly sampled from objects that do not appear in the image. The popular setting selects objects from the top half of the most frequently occurring categories in the entire image dataset, while ensuring that these objects are absent from the queried image. In the adversarial setting, all objects are first ranked based on their co-occurrence frequencies with the ground-truth objects, after which the top-k most frequent objects that are not present in the image are chosen.
\subsection{MME}
The official benchmark released by the original work~\cite{fu2025mmecomprehensiveevaluationbenchmark} was designed to evaluate model performance across 14 subtasks covering both perception and cognition. In this work, we use the \textbf{Existence, Count, Position, and Color} subset, which is most related to object and attribute hallucinations.
\subsection{FaithSocre}
FaithScore aims to evaluate free-form responses to open-ended questions. Specifically, it includes three steps: descriptive sub-sentence identification, atomic fact generation, and fact verification. In this work, we use LLaVA-v1.5-13B to achieve sub-sentence identification and LLaVA-1k collected by the original work~\cite{jing2024faithscorefinegrainedevaluationshallucinations} as the benchmark.

\subsection{GOAT-Bench}
Respecting Facebook's licenses on the memes, the GOAT-Bench dataset only contains the annotated text for the Facebook data, but not the hateful memes. Because of official restrictions from the Facebook Hateful Memes Challenge, we are currently unable to access these memes; therefore, we remove the hatefulness subtask from the benchmark.

\section{Additional experimental results on Qwen2-VL}
To further demonstrate the generality of our method across different model architectures, we additionally conduct the same experiments described above. Here, we report the performance of our proposed method on Qwen2-VL-7B~\cite{Qwen2VL} across different benchmarks, from which we observe consistent performance improvements to varying degrees.
\begin{table*}[]
\centering
\caption{Performance comparisons on CHAIR, POPE, and MME.}
\label{tab:qwen_hallu}
\resizebox{\textwidth}{!}{
\begin{tabular}{@{}clcccccccccc@{}}
\toprule
\multirow{2}{*}{Methods} & \multicolumn{3}{c}{CHAIR} & \multicolumn{2}{c}{POPE} & \multicolumn{2}{c}{MME} \\
\cmidrule(lr){2-4} \cmidrule(lr){5-6} \cmidrule(lr){7-8} \cmidrule(lr){9-10}
& Average $\downarrow$ & Recall $\uparrow$ & F1-score $\uparrow$ & Acc. $\uparrow$ & F1-score $\uparrow$ & Object-level $\uparrow$ & Attribute-level $\uparrow$ \\
\midrule
Baseline  & 14.91 & 68.4 & 75.84 & 88.99 & 88.60 & 225.00 & 303.33\\
Ours & \textbf{13.27} & \textbf{68.6} & \textbf{76.60} & \textbf{90.02} & \textbf{89.16} & \textbf{230.00} & \textbf{308.50} \\
\bottomrule
\end{tabular}}
\end{table*}

\begin{table*}[t]
\centering
\caption{Performance comparisons on the GOAT benchmark across different subcategories.}
\label{tab:qwen_goat}
\resizebox{\textwidth}{!}{
\begin{tabular}{l cc cc cc cc | cc}
\toprule
\multirow{2}{*}{Methods} & \multicolumn{2}{c}{Misogyny} & \multicolumn{2}{c}{Offensiveness} & \multicolumn{2}{c}{Sarcasm} & \multicolumn{2}{c}{Harmfulness} & \multicolumn{2}{c}{Average}\\
\cmidrule(lr){2-3} \cmidrule(lr){4-5} \cmidrule(lr){6-7} \cmidrule(lr){8-9} \cmidrule(lr){10-11}
& Acc. & F1 & Acc. & F1 & Acc. & F1 & Acc. & F1 & Acc. & F1\\
\midrule
Baseline & 71.50  & \textbf{69.39} & 62.72 & 43.58 & 52.80 & 15.37 & 59.08 & 30.40 & 61.53 & 39.69\\
Ours   & \textbf{71.65} & 69.17 & \textbf{63.89} & \textbf{50.27} & 52.80 & \textbf{47.76} & \textbf{61.34} & \textbf{42.19} & \textbf{62.42} & \textbf{52.35}\\
\bottomrule
\end{tabular}
}
\end{table*}

\begin{figure}[]
    \centering
    \includegraphics[width=0.7\columnwidth]{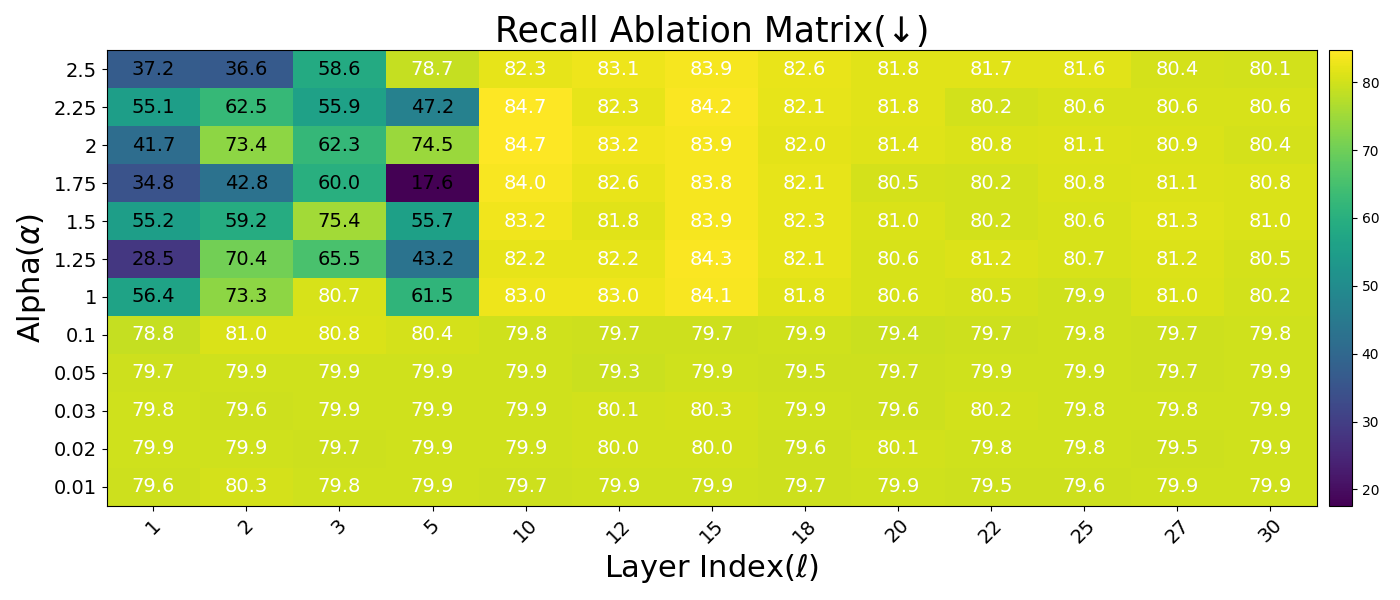}
    \caption{Ablation Recall matrix for OSGA steering strength($\alpha$) and layer $\ell$ on LLaVA-v1.5.}
    \label{fig:recall_matrix}
\end{figure}

\begin{figure}[]
    \centering
    \includegraphics[width=1.0\columnwidth]{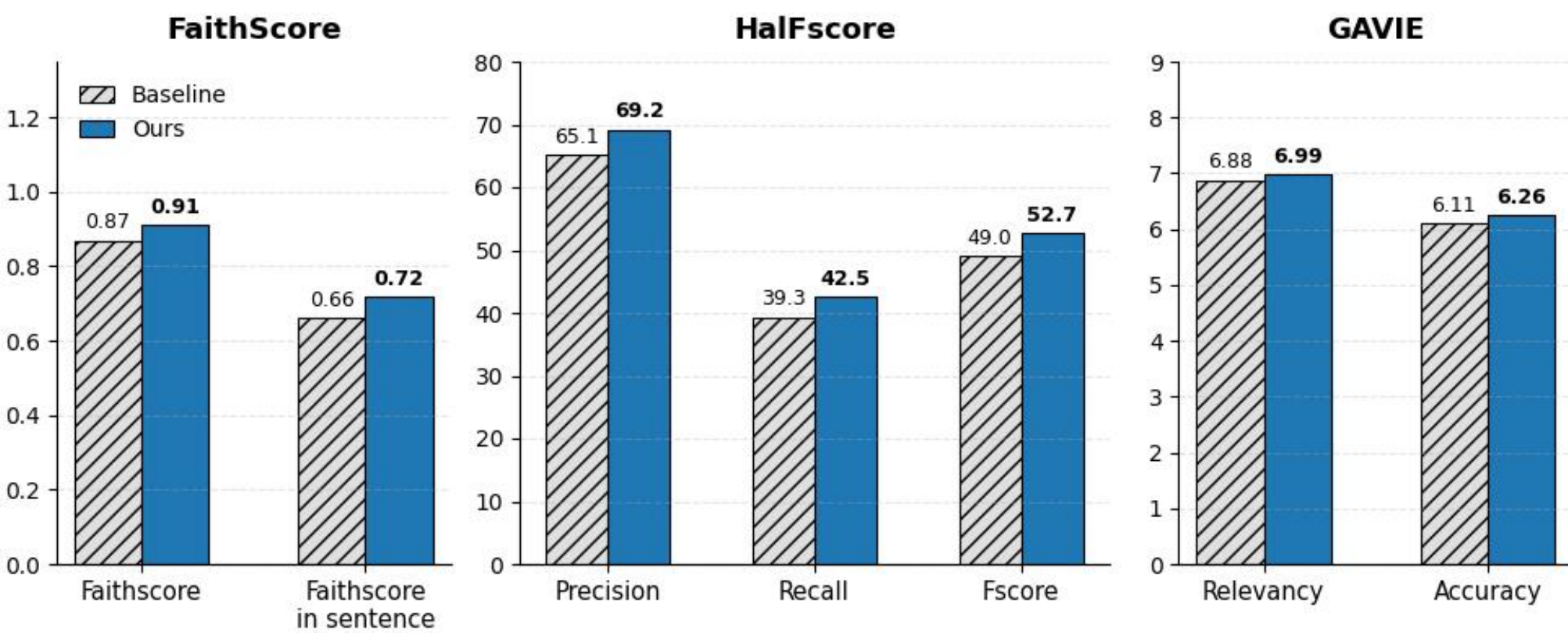}
    \caption{ process self-supervised.}
    \label{fig:qwen_benchmarks}
\end{figure}


\end{document}